\documentclass{article}
\usepackage{graphicx} 
\usepackage{xcolor}
\usepackage[hidelinks]{hyperref}
\usepackage{multirow}
\usepackage{graphicx}
\usepackage{booktabs}
\usepackage{tabularx}
\usepackage{makecell}
\usepackage[margin=1in]{geometry}
\usepackage{csvsimple}
\usepackage{amsmath}
\usepackage{graphicx}
\usepackage{booktabs}
\usepackage{subcaption}  
\usepackage{array}
\usepackage{url}        
\usepackage{authblk}
\usepackage[normalem]{ulem} 
\usepackage{pifont} 
\usepackage[labelfont=bf]{caption}

\newcommand{\smallbullet}{.}

\title{\huge {\bf{Surfer-H Meets Holo1}} \\ {Cost-Efficient Web Agent Powered by Open Weights}\\
\vspace{1cm}
\includegraphics[width=0.5\textwidth]{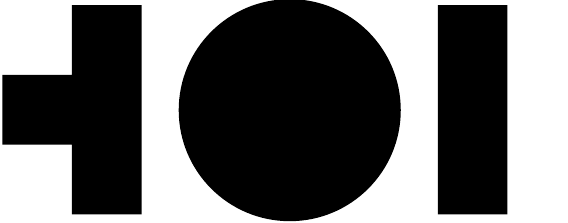}
}

\date{}  

\author{M\smallbullet~Andreux}
\author{B\smallbullet~Baldas Skuk}
\author{H\smallbullet~Benchekroun}
\author{E\smallbullet~Biré}
\author{A\smallbullet~Bonnet}
\author{R\smallbullet~Bordie}
\author{N\smallbullet~Bout}
\author{M\smallbullet~Brunel}
\author{P\smallbullet-L\smallbullet~Cedoz}
\author{A\smallbullet~Chassang}
\author{M\smallbullet~Chen}
\author{A\smallbullet D\smallbullet~Constantinou}
\author{A\smallbullet~d'Andigné} 
\author{H\smallbullet~de La Jonquière}
\author{A\smallbullet~Delfosse}
\author{L\smallbullet~Denoyer}
\author{A\smallbullet~Deprez}
\author{A\smallbullet~Derupti}
\author{M\smallbullet~Eickenberg}
\author{M\smallbullet~Federico}
\author{C\smallbullet~Kantor}
\author{X\smallbullet~Koegler}
\author{Y\smallbullet~Labbé}
\author{M\smallbullet C\smallbullet H\smallbullet~Lee}
\author{E\smallbullet~Le Jumeau de Kergaradec}
\author{A\smallbullet~Mahla} 
\author{A\smallbullet~Manevich}
\author{A\smallbullet~Maret}
\author{C\smallbullet~Masson}  
\author{R\smallbullet~Maurin}
\author{A\smallbullet~Mena}
\author{P\smallbullet~Modard}
\author{A\smallbullet~Moyal}
\author{A\smallbullet~Nguyen Kerbel}
\author{J\smallbullet~Revelle}
\author{M\smallbullet~L\smallbullet~Richter}
\author{M\smallbullet~Santos}
\author{L\smallbullet~Sifre}
\author{M\smallbullet~Theillard}
\author{M\smallbullet~Thibault}
\author{L\smallbullet~Thiry} 
\author{L\smallbullet~Tronchon} 
\author{N\smallbullet~Usunier} 
\author{T\smallbullet~Wu}

\affil{
\href{https://www.hcompany.ai/}{H Company} - Alphabetical order
}

\newcommand{\eg}{{\it e.g. }}
\newcommand{\ie}{{\it i.e. }}

\newcommand{\hfamilyname}{{Holo1}}
\newcommand{\hfamily}{\hfamilyname}  
\newcommand{\hthreeb}{{\hfamilyname-3B}} 
\newcommand{\hsevenb}{{\hfamilyname-7B}} 
\newcommand{\webclick}{{WebClick}}
\newcommand{\surferh}{{Surfer-H}}

\begin{document}

\maketitle

\begin{abstract}
We present \surferh, a cost-efficient web agent that integrates Vision-Language Models (VLM) to perform user-defined tasks on the web.
We pair it with \hfamily, a new open-weight collection of VLMs specialized in web navigation and information extraction.
\hfamily~was trained on carefully curated data sources, including open-access web content, synthetic examples, and self-produced agentic data. 
\hfamily~tops generalist User Interface (UI) benchmarks as well as our new web UI localization benchmark, \webclick.
When powered 
 by \hfamily, \surferh~achieves a 92.2\% state-of-the-art performance on WebVoyager, 
striking a Pareto-optimal balance between accuracy and cost-efficiency. To accelerate research advancement in agentic systems, we are open-sourcing both our \webclick~evaluation dataset and the \hfamily~model weights.
\end{abstract}

\section{Introduction}
Building AI agents requires designing systems capable of acting in and adapting to dynamic digital environments in real time. In this context, 
Large Language Models (LLMs) have made remarkable progress in reasoning and problem solving, rivaling or even surpassing human experts in domain-specific tasks \cite{GPT4-o-passes-bar-exam,shi2024languagemodelssolveolympiad}.
However, in their most fundamental form, LLMs are confined to a static, pre-trained world: they cannot act, verify, or access up-to-date information. For instance, they cannot answer questions about current events, book a restaurant table, or avoid hallucination \cite{rohrbach-etal-2018-object, xiao-wang-2021-hallucination}.

To circumvent their limitations, research has focused on enhancing LLMs with tool-use capabilities, enabling them to execute code snippets~\cite{google2024gemini,replit2022ghostwriter}, query Application Programming Interfaces (APIs)~\cite{ nakano2022webgptbrowserassistedquestionansweringhuman, schick2023toolformer},  
 or retrieve information at scale with multi-step reasoning  \cite{wei2022chain,yao2023reactsynergizingreasoningacting,openai2025deepresearch, perplexity2025deepsearch}.
 These systems, often referred to as \textit{agents}, extend LLMs into more capable virtual assistants \cite{xu2025theagentcompanybenchmarkingllmagents}.
 However, their real-world utility remains bounded by the available predefined tools and the engineering effort required to expand them \cite{koh2024visualwebarenaevaluatingmultimodalagents}.

Approaching this problem from another angle, computer use agents have recently emerged as a new paradigm in which agents interact with software directly through Graphical User Interfaces (GUIs) \cite{anthropic2024claude3, google2024projectmariner, webvoyager,lavague,  browser_use2024, openai2025operator, zhou2024webarena}, \ie using the same interface humans are presented with. This approach avoids relying on custom integrations or APIs, opening the door to more adaptable general-purpose agents with higher potential and broader real-world utility.

Here we present \surferh\footnote{\url{https://www.surferh.com}}, a visual web retrieval agent designed to be easily trained through reinforcement learning techniques.
\surferh~comprises three main modules: a policy, a localizer, and a validator, which act in sequence (see Section~\ref{sec:surferh}).
These modules are compatible with any VLM capable of proposing and evaluating actions. Our agent only uses screenshots from websites and does not require the Document Object Model (DOM) or the accessibility tree of the websites.
To deliver the best cost-performance ratio, we introduce \hfamilyname, a family of lightweight VLMs specialized in taking and evaluating actions and localizing UI elements.
\hfamily~was trained on carefully curated data sources, including open-access web content, synthetic examples, and self-produced agentic data.
\hfamily~is publicly available on Hugging Face\footnote{ \url{https://huggingface.co/collections/Hcompany/holo1-683dd1eece7eb077b96d0cbd}}.

Localization, the ability to identify the precise coordinates of User Interface~(UI) components in a screenshot, is a core capability for effective web navigation and interaction. Existing benchmarks, such as Screenspot~\cite{screenspot, screenspot-pro, screenspot-v2} and GroundUI~\cite{rawles2025androidworld}, primarily focus on general UI localization across apps and platforms, but are not tailored to the unique challenges of web-based environments. These include complex, dynamic components like calendars and nested menus that frequently appear during web navigation. To fill this gap, we introduce \webclick, a new benchmark specifically designed for web localizers, which we make publicly available on Hugging Face\footnote{\url{https://huggingface.co/datasets/Hcompany/WebClick}}. 
As detailed in Section~\ref{sec:localizer},
\webclick~features specialized UI elements representative of the modern web, extracted from human-annotated data and agentic on-policy interactions. 
Our results show that \hfamily~models excel at the aforementioned benchmarks.

We evaluate \surferh~on WebVoyager~\cite{webvoyager}, a mainstream benchmark for web retrieval, and compare the performance of \hfamily~models against external baselines in Section~\ref{sec:results}. Our results show that the accuracy and cost-efficiency of the \hfamily~models enable \surferh~to achieve state-of-the-art performance at Pareto optimality.

\section{\surferh} \label{sec:surferh}

\begin{figure}[t!]
    \centering
    \includegraphics[width=0.99\textwidth, keepaspectratio]{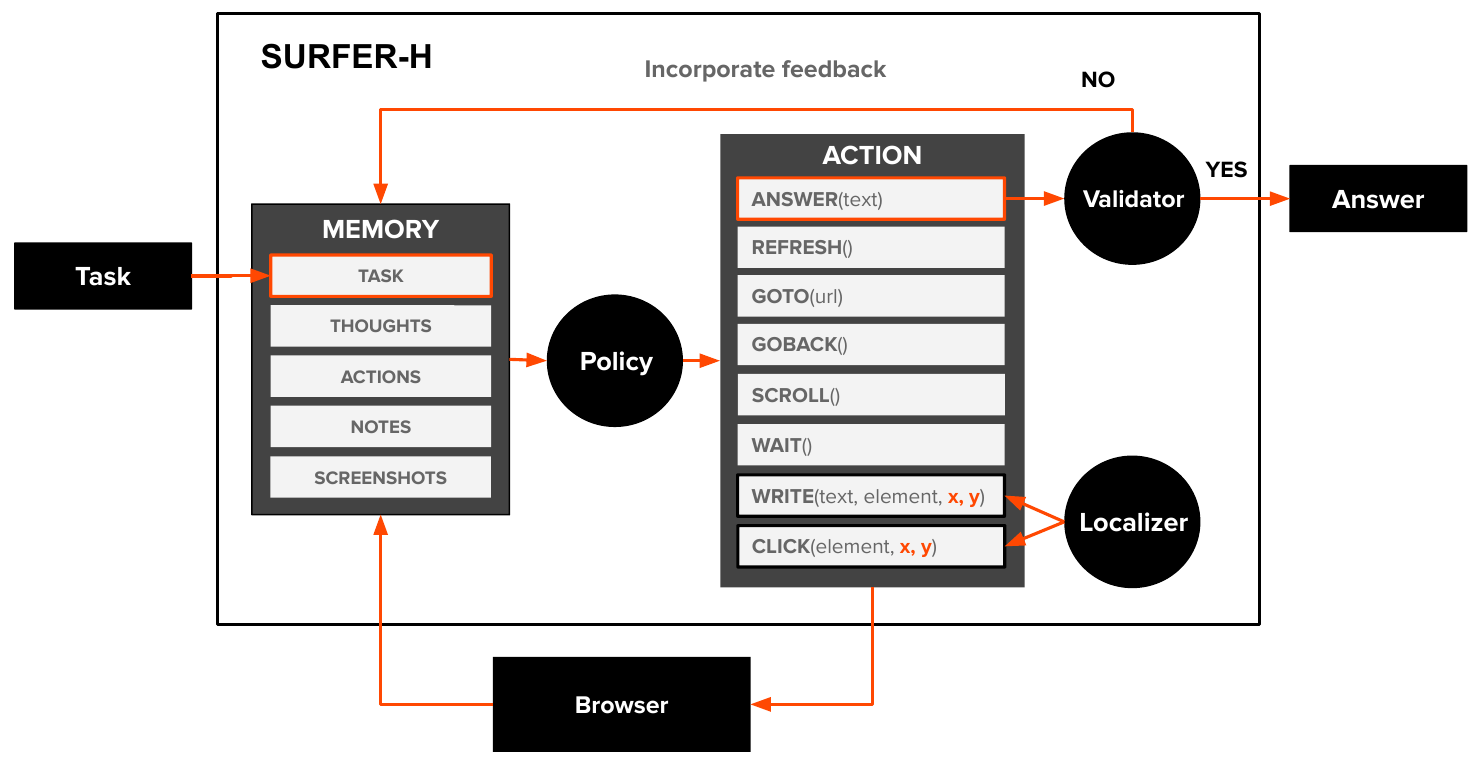}
    \caption{
   \textbf{\surferh} operates via screenshots and a limited action set. It maintains an internal memory with the task, recent screenshots, and thought history. The policy generates thoughts and selects the next action. If necessary, the localizer refines the coordinates for clicks or typing. The validator filters answers, gives feedback, and decides whether a task is complete.}
    \label{fig:surfer-h}
\end{figure}

As displayed in Figure~\ref{fig:surfer-h},
\surferh~relies on three trainable modules: a policy, a localizer, and a validator.
The policy proposes actions that are executed sequentially.
Web actions are executed in a browser by simulating human-like interactions.
If the policy generates an action that requires interacting with a specific element on a webpage, such as a like button, it generates a textual description of the element, and the localizer provides its 2D coordinates.
The policy can also decide that the task is complete and generate a textual answer via a dedicated action.
When the policy emits an answer, it is passed through the validator.
The validator generates feedback about the answer and decides whether it is suitable for the user.
If the answer is valid, it is returned to the user.
Otherwise, the feedback is incorporated to the agent's memory, and the agent continues its execution until either completion or reaching a time or cost budget. 
We describe the three modules and the role of the memory in greater detail below.

\paragraph{Action Space and Policy} Our agent is equipped with a small action space: it can click on or type text into particular web components, scroll up and down, wait for a page to load, refresh the page, go to a given URL, go back, or return an answer. For a given task and memory, \surferh~decides which action to execute using its policy, a specialized VLM. Each action is preceded by a thought and, if deemed necessary by the agent, note-taking. The thoughts, notes, and actions are stored in natural language so they can be easily interpreted. They are created using chain-of-thought prompting \cite{wei2022chain} and structured generation \cite{dong2024xgrammar}.

\paragraph{Memory} 
The past actions are stored in the agent's internal memory, along with the  most recent screenshots, thoughts, notes (\ie information gathered by \surferh~throughout the episode), and the current browser view.
\surferh~maintains and iteratively updates its internal memory, and uses it to produce one action at each timestep. 

\paragraph{Localizer}
For click and write actions requiring an element (\eg a button or search bar), a localizer identifies and integrates the element’s coordinates into the action. The localizer is a specialized UI model optimized for coordinate generation.
 
\paragraph{Validator}
If \surferh~believes it has successfully completed the user request, it will call the answer action and produce a textual answer with supporting screenshots. The validator will then be summoned to review the generated answer for approval. If approved, the agent run is terminated and the answer is returned to the user. Otherwise, \surferh~will gather feedback from the validator in its notepad and continue browsing.
For each attempt, we set a maximum number of steps, at which point it is forced to produce an answer.

\paragraph{Models}
\surferh~modules can be powered by generalist foundation models or fine-tuned specialists. 
Note that the different modules can be served through different VLMs, or may rely on a single model using different prompting strategies, see Section~\ref{sec:results}.
In Section~\ref{sec:modules}, we describe how we trained \hfamily~to be used in any of these modules to deliver the best performance.

\section{Training the \hfamily~Family } \label{sec:modules}

\subsection{Overview}

Training \surferh~means training its constituent modules to perform their required tasks optimally within a web-browsing environment.
Our goal in training is therefore to imbue our models, the \hfamily~family, with a deep understanding of complex information on webpages and a precise state-to-action mapping.
We achieve this using a large-scale mixture of diverse datasets designed to capture the breadth and complexity of the modern web.
This mixture spans real-world web pages, synthetic UIs, document visualizations, and agent-based behavioral traces.
In this way, we encourage models to develop actionable understanding beyond surface-level recognition and enable generalization across a broad range of web interfaces.

\begin{table}[h]
\centering
\caption{\textbf{Dataset Distribution} with mixture group breakdown (tokens in billions).}
\begin{tabular}{@{}llrr@{}}
\toprule
\textbf{Dataset Group} & \textbf{Mixture Group} & \textbf{Tokens (B)} & \textbf{Percentage} \\
\midrule
\multirow{5}{*}{\textbf{GUI Grounding}} 
  & WebCrawl                     & 12.19                & 38.76  \\
  & Open-source datasets         & 3.42                 & 10.87  \\
  & WebSynthetic                 & 0.37                 & 1.17  \\
  & \textbf{Total}               &  \textbf{15.98}      & \textbf{50.79} \\
\midrule
\multirow{4}{*}{\textbf{Complex Visual Understanding} } 
  & Coordinate Validation         & 2.70                 & 8.59 \\
  & UI Extraction           & 5.93                 & 18.86  \\
  & VQA                          & 1.52                 & 4.84 \\
  & \textbf{Total}               & \textbf{10.16}       & \textbf{32.28} \\
\midrule
\multirow{3}{*}{\textbf{Behavior Learning} } 
  & Policy                      & 4.87                  & 15.48 \\
  & Validator                   & 0.46                  & 1.45  \\
  & \textbf{Total}              & \textbf{5.32}         & \textbf{16.93} \\
\midrule
\multicolumn{2}{l}{\textbf{Grand Total}} & \textbf{31.46} & \textbf{100.00} \\
\bottomrule
\end{tabular}
\label{tab:curriculum_distribution}
\end{table}
\subsection{Data Composition Summary}
The different elements of the training mixture described in Table~\ref{tab:curriculum_distribution} reflect the various model capabilities that are expressed in the localizer, policy and validator modules. 

The foundation of the training mixture is GUI grounding data and is composed of web-crawled and synthetic pages, labeled for the detection of UI elements based on visual cues, as well as open-source datasets. This data source forms 50.79\% of the total tokens and consists mainly of proprietary data, distinguishing our models from those trained solely on public datasets.

We enhance our overall dataset mixture with data for Complex Visual Understanding, which covers tasks such as assessing the localizer outputs, extracting the interactable elements from web pages, and Visual Question Answering (VQA). This more specialized data amounts to 32.28\% of our training mixture.

The third tranche of the mixture comprises data collected from our agents in action, enabling our models to learn from past behavior, and amounts to 16.93\% of the overall training data. It contains a set of action datasets based on past successful agent traces, which represents 15.48\% of training tokens.
It allows the model to learn complex patterns of memory management, paired thinking and action generation, and an understanding of states in the context of a broader task. Additionally, the mixture also contains examples of evaluation of an agent's answer against a task based on textual and visual evidence. This represents 1.45\% of the tokens used during training.

\subsection{GUI Grounding}
The localizer is essential for bridging the visual and action spaces: it enables the agent to determine the coordinates of an interface element selected for interaction based on visual cues. To be effective, this capability must generalize across the vast diversity of web pages, which vary widely in language, layout, visual style, content density, and interactive complexity.

\paragraph{WebCrawl Dataset}
To address this, we constructed the WebCrawl dataset, a large-scale collection of web pages sampled from the public Internet. The HTML content of each page was parsed, and elements that allow for interaction (\eg click, text input, selection) were extracted. These elements are then mapped to an intent, which ranges from simple text content to high-level intents (e.g., ``submit search query" or ``open user settings") that require reasoning about UI functionality beyond literal appearance. These abstract action descriptions are synthetically generated using frontier models.

We collected click interaction data from \(4\) million web pages, amounting to \(89\) million clicks in total. Furthermore, we rely on open datasets such as OS-Atlas \cite{screenspot-v2} to complement and diversify the mixture. Each sample pairs an image and an intent with precise click coordinates, which are used as labels. 

\paragraph{WebSynthetic Dataset}

We strategically augment the generalist mixture with proprietary synthetic datasets to address known challenges in UI grounding. These carefully crafted resources include the following:
\begin{itemize}
\item Custom-developed websites with calendars and relevant intents, addressing a task known to challenge web agents.
\item Synthetic tables with relational data structures, targeting a known weakness in current models that struggle to properly interpret tabular information spread across multiple rows and columns.
\item A synthetic dataset focused on icon interpretation on the web, enabling improved recognition and functional understanding of ambiguous UI elements that standard datasets consistently misclassify.
\end{itemize}
These proprietary synthetic sources specifically target failure cases observed in conventional models, serving as adversarial training data that significantly enhances model robustness beyond what is possible with standard datasets.

\subsection{Complex Visual Understanding}

More complex visual understanding capabilities are introduced in the models by training on specialized datasets which enable grounding analysis, precise information extraction, and Visual Question Answering~(VQA).

\paragraph{Coordinate Validation Data} We introduce a novel dataset for the judgment of a grounding proposition. Here, given a triplet consisting of an image, intent, and coordinates, models predict whether the click action aligns with the stated intent. This leverages a Set-of-Marks \cite{yang2023setofmarkpromptingunleashesextraordinary} approach to highlight the area of interest and trains models to evaluate the match between textual intent and visual targets. This dataset contains more than 5 million triplets for Coordinate Validation.

\paragraph{UI Extraction Data}
Models are trained to exhaustively extract every clickable, selectable, or inputtable element on a page. Given a screenshot, the model outputs the (a) location and (b) label of each interactable element. This goes beyond standard Optical Character Recognition (OCR) tasks by emphasizing cues that signal interactivity, such as affordances in fonts, frames, and styling, and encourages models to be exhaustive and non-redundant in extraction. Our UI Extraction dataset contains close to \(7\) million pages.

\paragraph{VQA}
We use common datasets for visual understanding and question answering; ingesting, remapping, and filtering them to extract their most valuable components. We focus on chart, table, and document understanding, and parts of Cauldron by Huggingface \cite{laurençon2024matters}, totaling 600,000 images. We enrich the training mix with internal datasets tailored for complex visual understanding. These datasets focus on interpreting charts, dashboards, tables, and dense reports, enabling the extraction of numerical, relational, and scientific information. Together, these datasets contain 150M tokens and 300,000 images, giving our model an advantage over those trained on public visual data alone.

\subsection{Behavior Learning on Multimodal Traces}
\label{sec-bc}
\paragraph{Multimodal Traces Data} Crucially, our training dataset includes multimodal traces from agent executions. These elements allow the model to bridge the gap between vision and action by representing action messages grounded in visual inputs. They also encourage memory understanding and planning by learning actions as a function of past observations and actions. Finally, they represent the grounding of actions in a thinking pattern by jointly predicting thought, notes and action pairs:
\begin{equation}
\label{eq:state-output}
\left(\text{thought}_{t+1}, \text{notes}_{t+1}, \text{action}_{t+1}\right)
\sim
\pi \left(\text{task}, \{\text{thought}_k, \text{action}_k, \text{notes}_k, \text{screenshot}_{t-3 < k} ~|~ k\leq t \}\right).
\end{equation}

These sequences train for the exact policy VLM used in the \surferh\ logic, and represent the learning of a mapping between memory and action defined in Equation~\ref{eq:state-output}.

\paragraph{Offline Reinforcement Learning}
These trajectories teach models to behave as agents: reasoning over long contexts, understanding goals, and predicting the next action based on task history. Following a Filtered Behavioral Cloning (FBC) approach, only successful traces are retained in the final mixture. This component is essential for upgrading the model from passive understanding to active, interactive web navigation. Each agent execution contains up to 30 policy steps, resulting in a large amount of training tokens.

Agent trajectories for this dataset were generated using two task corpora. The first is WebVoyager~\cite{webvoyager}, which comprises 643 tasks on 15 common websites, mostly consumer-facing. The second is a new corpus we generated, WebVoyagerExtended, with 15,000 tasks spanning 330 websites. To construct it, we identified websites similar in function, features, and design to those in WebVoyager, then synthetically generated tasks mirroring WebVoyager’s style.

The introduction of the former traces demonstrates the self-play and self-learning capability of our system, \ie our agent's ability to improve from past executions.
The latter extend the robustness of the policy capabilities of the model by promoting data diversity.
We investigate the relative impact of self-learning and learning on broader tasks in Section~\ref{sec:generalization}.

\subsection{Feedback and Validation Learning}
\paragraph{Validation Data Format}
The ability of \surferh~to inspect and validate a proposed answer before submission is crucial. This is formalized by a function $V$ (Equation\eqref{eq:validator-eqn}), which takes as input the task, textual answer, and supporting screenshots. The validator outputs a boolean indicating task success, along with an explanation justifying the decision. This explanation guides subsequent attempts by \surferh.
\begin{equation}
\label{eq:validator-eqn}
(\text{success}, \text{explanation})
\sim
V \left(\text{task}, \text{answer}, \lbrace\text{screenshot}_{t-3 < k \leq t} \rbrace\right). 
\end{equation}
\paragraph{Learning from Past Validations}
Similar to Section~\ref{sec-bc}, we generate more than 1 million validation input and output pairs, based on real agent executions and answers, on the aforementioned websites. The proposed validation inputs therefore represent realistic agent answers, together with evidence in the form of screenshots gathered on agent trajectories. The output explanations and validation Boolean are generated by frontier VLMs, prompted to evaluate the validity of the answer, and the grounding of the answer in provided screenshots.

\subsection{Training Strategy}

\hfamily~models are trained using a mixture of text completion and tool call samples, encouraging them to follow instructions, leverage context, and predict actionable outputs in both passive (extraction) and active (interaction) tasks. While data is organized around layered capabilities, we transform each sample into a chat-like example with system, user and assistant messages, with one or multiple images per input. Consequently, the training dataset is effectively a multi-task and multi-modal chat dataset mixture.

Regarding the \hfamily~models themselves, we start from Qwen 2.5-VL-Instruct \cite{bai2025qwen25vltechnicalreport} weights that we fine-tune using our proprietary training codebase. Instead of one model per module, each \hfamily~model is trained on the entire dataset to cover all module capabilities (policy, localizer, validator), as well as other standard VLM capabilities. In doing so, we allow our models to handle both low-level (localization) and high-level (policy, validation) operations with variable model size. This allows us to measure and control the cost-effectiveness of our agents.

We used the ToxiGen dataset~\cite{hartvigsen2022toxigenlargescalemachinegenerateddataset} to evaluate the toxicity of the \hfamily~model outputs. We found that only 2.1\% and 1.5\% of the responses were flagged, for \hthreeb~and \hsevenb~respectively. As a reference, we found that Qwen2.5-VL 3B and 7B score 3.7\% and 0.5\%, respectively. This suggests that the safety of the initial models was well preserved by the training procedure.

\section{\hfamily~Localization Skills} \label{sec:localizer}
\subsection{Overview}

Localization is a key skill for the real-world utility of our VLMs. The ability to identify precise coordinates on a UI determines the success of a click or write action and thus the capacity to complete a task. To assess this capability, we evaluated our \hfamily~models on several established localization benchmarks, including Screenspot \cite{screenspot}, Screenspot-V2 \cite{screenspot-v2}, Screenspot-Pro \cite{screenspot-pro}, GroundUI-Web \cite{rawles2025androidworld}, and our own newly introduced benchmark, \webclick, described in Section~\ref{webclick}.
For comparison, we also evaluated current state-of-the-art~VLMs: UI-TARS \cite{qin2025ui}, and the Qwen-2.5-VL and UGround-V1 families \cite{bai2025qwen25vltechnicalreport, gou2025uground}.

\subsection{\webclick : A Specialized Web Localization Benchmark} \label{webclick}
 We introduce a new web localization benchmark called \webclick~to accurately measure and track model performance on localization and, by extension, web interaction capabilities.
This benchmark follows a screenspot-like format, wherein each datapoint includes a web screenshot, an instruction, and a bounding box that marks the interactive element to be clicked in order to complete the task. The VLM being tested receives a screenshot and instruction, and is asked to respond with interaction coordinates; correct answers are those that fall within the bounding box.

We carefully curated this benchmark dataset from three sources: (1) data collected by our agents while attempting to solve WebVoyager tasks \cite{webvoyager}, (2) human interactions with the Web during everyday tasks, and (3) human interactions with calendar interfaces. The calendar data, a subset of human interactions with the Web, was deliberately isolated and developed into its own dataset, as we identified calendar navigation as a setting in which many contemporary VLMs underperform. Moreover, accurately using and understanding calendars is particularly important for enhancing the practical utility of our agents. Through manual curation, we ensured this benchmark includes challenges commonly observed as points of failure in state-of-the-art models, such as understanding UI conventions or combining textual instruction with visual reasoning. Calendar tasks can be particularly difficult, requiring models to interpret structural elements and account for regional variations in date formats. In total, the benchmark contains 1,639 screenshots from over 100 websites. The benchmark is publicly released under the Apache-2 license, and can be found at~\url{https://huggingface.co/datasets/Hcompany/WebClick}.

\begin{table}[h!]
\centering
\caption{\textbf{Click Accuracy (\%)} across models and benchmarks.}
\renewcommand{\arraystretch}{1.2}
\resizebox{\textwidth}{!}{
\begin{tabular}{@{}lccccccccc@{}}
\toprule
\textbf{Model} 
& \multicolumn{3}{c}{\textbf{Screenspot}} 
& \textbf{GroundUI} 
& \multicolumn{3}{c}{\textbf{\webclick}~(ours)} 
& \textbf{Avg} \\
\cmidrule(lr){2-4}
\cmidrule(lr){6-8}
& v1 \cite{screenspot} & v2 \cite{screenspot-v2} & Pro \cite{screenspot-pro} & Web \cite{rawles2025androidworld} & agent & calendar & human & \\
\midrule
Qwen2.5-VL-3B-Instruct \cite{bai2025qwen25vltechnicalreport} & 82.78 & 84.34 & 7.91  & 70.50 & 76.26 & 51.70 & 85.07 & 65.51  \\
UGround-V1-2B \cite{gou2025uground}          & 77.12 & 79.31 & 21.32 & 78.60 & \textbf{84.41} & 50.76 & 78.50 & 67.15  \\
UI-TARS-2B \cite{qin2025ui}             & 66.82 & 69.39 & 16.38 & \textbf{80.75} & 78.68 & 42.05 & 70.33 & 60.63\\
\hthreeb~(ours)             & \textbf{85.93} & \textbf{88.91} & \textbf{23.66} & 74.75 & 83.02 & \textbf{65.91} & \textbf{88.80} & \textbf{73.55} \\

\midrule

Qwen2.5-VL-7B-Instruct \cite{bai2025qwen25vltechnicalreport} & 85.53 & 88.04 & 10.12 & 78.75 & 78.47 & 59.09 & 85.22 & 69.32 \\
UGround-V1-7B \cite{gou2025uground}          & 85.69 & 84.26 & \textbf{30.93} & \textbf{82.70} & \textbf{92.37} & 68.75 & 84.84 &  75.65 \\
UI-TARS-7B \cite{qin2025ui}         & 84.20 & 86.70 & 23.53 & 81.00 & 90.47 & 63.45 & 87.03 & 73.77\\
\hsevenb~(ours)             & \textbf{87.42} & \textbf{89.85} & 26.06 & 78.50 & 89.77 & \textbf{72.92} & \textbf{88.80} &  \textbf{76.19}\\
\bottomrule
\end{tabular}
}
\label{tab:click_accuracy}
\end{table}

\begin{figure}[h!]
    \centering
    \includegraphics[width=0.99
    \linewidth]{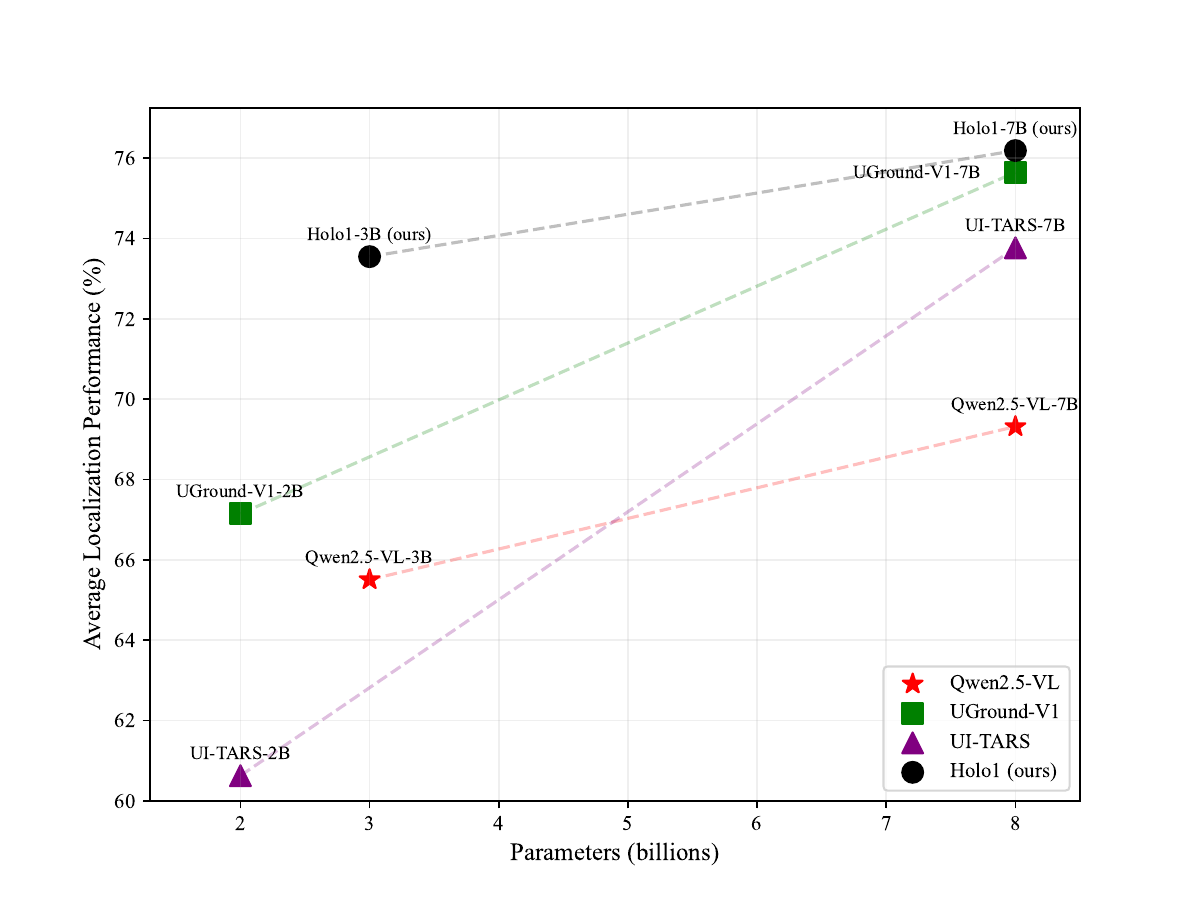}
    \caption{{\bf \hfamily~as Localizer}: comparison against competitors, for external and internal benchmarks. \hfamily~models reach state-of-the-art average localization performance at all model scales.}
    \label{fig:localization-pareto}
\end{figure}

\subsection{Localization Benchmarks Results}
Overall, we find that our \hfamily~model family outperforms state-of-the-art models of a similar size, as reported in Table~\ref{tab:click_accuracy} and Figure~\ref{fig:localization-pareto}. Both \hthreeb~and \hsevenb~achieve the highest average localization performance for models of their size, scoring 73.55\% and 76.16\%, respectively. The \hthreeb~model demonstrates strong performance across both public and internal benchmarks. For example, we observe a significant improvement over Qwen2.5-VL-3B, UGround-V1-2B and UI-TARS-2B on the human-based elements of \webclick, as well as on all Screenspot variants. 

Furthermore, \hthreeb~not only achieves the highest average localization performance of the 2B and 3B models, but also outperforms the larger Qwen2.5-VL-7B by 4.23 percentage points, and is competitive with UGround-V1-7B, with less than 0.5 percentage points separating their scores.
The \hsevenb~variant improves upon the performance of the \hthreeb~model and maintains its position ahead of its competitors, validating the scalability of the \hfamily~training framework.  Across two of the three Screenspot benchmarks, and two of the three \webclick~datasets, it surpasses UGround-V1-7B.
Despite falling slightly behind its competitors on GroundUI, with a score of 78.50\%, it still achieves the highest average score of 76.19\%.
In summary, our \hfamily~models excel in localization, which stands them in good stead for their use in \surferh, which we explore and evaluate in Section~\ref{sec:results}.

\section{Surfing WebVoyager} \label{sec:results}
\subsection{Methodology}

\paragraph{WebVoyager Setup}
We evaluate \surferh~and external competitors on the WebVoyager benchmark \cite{webvoyager}, using all 643 tasks from 10 different websites.
For date-dependent tasks, we adjust the original dates so that they are always in the future relative to when the benchmark is executed, preventing invalid lookups, such as attempting to book a cruise that set sail the month before last. 
Success is computed as the majority vote from three samples of GPT-4o.

\paragraph{Baselines}
We use BrowserUse~\cite{ browseruse-tech-report-2024,browser_use2024}, Project Mariner~\cite{google2024projectmariner}, and OpenAI Operator~\cite{openai2025operator} as external baselines.
We use reported numbers for all of these baselines.
It should be noted that these reported scores were computed at a different time, with slightly different websites, website contents, and evaluation functions.
As baselines for the \hfamily~family within \surferh, we use the GPT-4~\cite{gpt4} and Qwen2.5~\cite{bai2025qwen25vltechnicalreport} model suites, along with Gemini-Flash-2.0~\cite{google2024gemini}.

\paragraph{\surferh~Configuration}
For each task, \surferh~is permitted up to 30 steps to produce an answer. If no valid response is generated within this limit, a final forced response is issued. When a response is rejected by the validator, \surferh~attempts to retry, with a maximum of 10 attempts per task.

\begin{figure}[t!]
    \centering
    \includegraphics[width=0.99\textwidth]{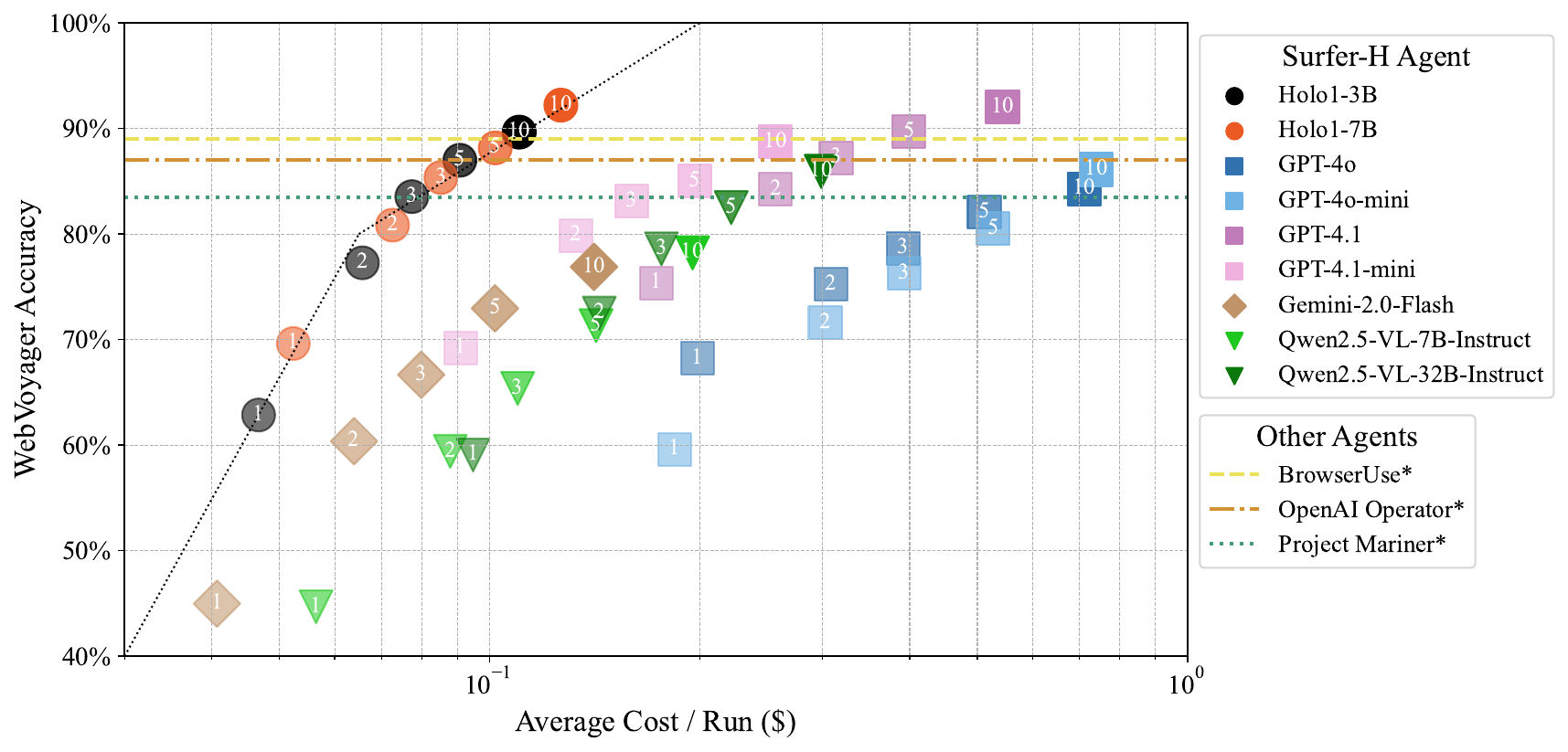}
         \caption{\textbf{Pareto-Optimality of \surferh+\hfamily}. \surferh~success is plotted against cost for varying maximum allowed attempts before the agent must respond, and different underlying policy modules (\hfamily, GPT, Gemini, or Qwen2.5-VL). For BrowserUse~\cite{browseruse-tech-report-2024}, OpenAI Operator~\cite{openai2025operator} and Project Mariner~\cite{google2024projectmariner}, we use reported numbers (* superscript).
         \surferh~powered by \hfamily~models reaches state-of-the-art performance while being the most cost-efficient.
         }
    \label{fig:pareto-agents-retries}
\end{figure}

\begin{table}[h!]
\centering
\caption{\textbf{Inference Cost Per Model Type}, using official external providers costs and internal cost estimation for the \hfamily~and Qwen2.5-VL models.
}
\renewcommand{\arraystretch}{1}
\begin{tabular}{lccc}
\toprule
\textbf{Model} &\makecell{\textbf{Cost(\$) /} \\\textbf{M input tokens}} & 
\makecell{\textbf{Cost(\$) /} \\\textbf{M output tokens}} & \makecell{\textbf{Token count /} \\ \textbf{1200x1200 image}}
\\
\midrule
GPT-4o~\cite{gpt4o2025cost} & 2.5 & 10 & 772 \\
GPT-4o-mini~\cite{gpt4o-mini2025cost} & 0.15 & 0.6 & 25508 \\
GPT-4.1~\cite{gpt41-2025cost} & 2 & 8 & 772 \\
GPT-4.1-mini~\cite{gpt41-mini-2025cost} & 0.4 & 1.6 & 2348 \\
Gemini-2.0-Flash~\cite{gemini2025cost} & 0.1 & 0.4 & 1290 \\
\hthreeb & 0.1 & 0.4 & 1280 \\
\hsevenb/Qwen2.5-VL-7B-Instruct & 0.15 & 0.6 & 1280 \\
Qwen2.5-VL-32B-Instruct & 0.5 & 2 & 1280 \\
\bottomrule
\end{tabular}

\label{tab:inference-cost}
\end{table}

\begin{table}[b!]
\centering
\caption{\textbf{Impact of the Training Mixtures}.
While the maximum performance is achieved with in-domain knowledge (\hsevenb), web navigation benefits from cross-domain exploration (\hsevenb-WVE).
}
\begin{tabular}{@{}lccccccccc@{}}
\toprule
\textbf{Policy Training Tasks} & \bf Policy Module &\makecell{\textbf{WebVoyager} \\ \textbf{Accuracy (\%)}} \\
\midrule
None 
&Qwen2.5-VL-7B-Instruct& 78.2  \\
WebVoyagerExtended only                 &\hsevenb-WVE& 87.7\\
WebVoyager + WebVoyagerExtended   &\hsevenb& 92.2 \\
\bottomrule
\end{tabular}

\label{tab:gen}
\end{table}

\paragraph{Metrics} We track the average WebVoyager success accuracy as a function of the average cost per run (\ie across all 643 tasks) in U.S. dollars. This cost is estimated based on the total usage of agent modules throughout the completion of the task. For modules powered by external APIs, we use the official pricing of the providers. For modules implemented using \hfamily~or Qwen2.5-VL models, we rely on internal estimates of inference cost based on model size. The per token prices we use are available in Table~\ref{tab:inference-cost}, along with the typical per image token counts. Both are needed for accurate and meaningful comparisons, especially as external providers have very distinct pricing models.
Overall, this cost reflects both task difficulty (harder tasks require more steps and attempts) and model complexity (larger models incur higher inference costs). Metrics are reported as a function of the maximum number of attempts executed.

\begin{table}
    \centering
    \caption{\textbf{\surferh~WebVoyager Performance} for various modules configurations and against external agents. The superscript * indicates reported numbers. All others are computed internally.}\label{table:pareto-numbers}
    \begin{tabular}{lcccccc}
\toprule
\textbf{Agent} & \textbf{Policy} & \textbf{Localizer} & \textbf{Validator} & \textbf{Attempts} 
& \makecell{
\textbf{Accuracy} \\ \textbf{(\%)}} & \makecell{\textbf{Cost} \\ \textbf{(\$/task)}} \\
\midrule
\multirow{40}{*}{\surferh} 

& \multirow{4}{*}{GPT-4o} & \multirow{4}{*}{Holo1-3B} & \multirow{4}{*}{GPT-4o} & 1 & 68.2 & 0.20 \\
&                         &                           &                          & 2 & 75.2 & 0.31 \\
&                         &                           &                          & 5 & 82.1 & 0.51 \\
&                         &                           &                          & 10 & {84.3} & {0.71} \\
\cline{2-7}

& \multirow{4}{*}{GPT-4o-mini} & \multirow{4}{*}{Holo1-3B} & \multirow{4}{*}{GPT-4o} & 1 & 59.6 & 0.18 \\
&                              &                           &                          & 2 & 71.6 & 0.30 \\
&                              &                           &                          & 5 & 80.5 & 0.53 \\
&                              &                           &                          & 10 & {86.1} & {0.74} \\
\cline{2-7}

& \multirow{4}{*}{GPT-4.1} & \multirow{4}{*}{Holo1-3B} & \multirow{4}{*}{GPT-4o} & 1 & 75.4 & 0.17 \\
&                          &                           &                          & 2 & 84.2 & 0.26 \\
&                          &                           &                          & 5 & 89.7 & 0.40 \\
&                          &                           &                          & 10 & {92.0} & {0.54} \\
\cline{2-7}

& \multirow{4}{*}{GPT-4.1-mini} & \multirow{4}{*}{Holo1-3B} & \multirow{4}{*}{GPT-4o} & 1 & 69.2 & 0.09 \\
&                               &                           &                          & 2 & 79.8 & 0.13 \\
&                               &                           &                          & 5 & 85.0 & 0.20 \\
&                               &                           &                          & 10 & {88.8} & {0.26} \\
\cline{2-7}

& \multirow{4}{*}{Gemini-2.0-Flash} & \multirow{4}{*}{Holo1-3B} & \multirow{4}{*}{GPT-4o} & 1 & 45.0 & 0.04 \\
&                               &                           &                          & 2 & 60.4 & 0.06 \\
&                               &                           &                          & 5 & 73.0 & 0.10 \\
&                               &                           &                          & 10 & {76.9} & {0.14} \\
\cline{2-7}

& \multirow{4}{*}{Qwen2.5-VL-7B-Instruct} & \multirow{4}{*}{Holo1-3B} & \multirow{4}{*}{GPT-4o} & 1 & 44.7 & 0.06 \\
&                               &                           &                          & 2 & 59.4 & 0.09 \\
&                               &                           &                          & 5 & 71.3 & 0.14 \\
&                               &                           &                          & 10 & {78.2} & {0.20} \\
\cline{2-7}

& \multirow{4}{*}{Qwen2.5-VL-32B-Instruct} & \multirow{4}{*}{Holo1-3B} & \multirow{4}{*}{GPT-4o} & 1 & 59.1 & 0.09 \\
&                               &                           &                          & 2 & 72.5 & 0.14 \\
&                               &                           &                          & 5 & 82.5 & 0.22 \\
&                               &                           &                          & 10 & {85.9} & {0.30} \\
\cline{2-7}

& \multirow{4}{*}{Holo1-3B} & \multirow{4}{*}{Holo1-3B} & \multirow{4}{*}{GPT-4o} & 1 & 62.9 & 0.05 \\
&                           &                           &                          & 2 & 77.3 & 0.07 \\
&                           &                           &                          & 5 & 87.0 & 0.09 \\
&                           &                           &                          & 10 & {89.7} & {0.11} \\
\cline{4-7}

&                           &                           & \multirow{4}{*}{Holo1-3B} & 1 & 50.7 & 0.01 \\
&                           &                           &                          & 2 & 63.8 & 0.02 \\
&                           &                           &                          & 5 & 70.8 & 0.03 \\
&                           &                           &                          & 10 & {73.2} & {0.04} \\
\cline{2-7}

& \multirow{4}{*}{Holo1-7B} & \multirow{4}{*}{Holo1-7B} & \multirow{4}{*}{GPT-4o} & 1 & 69.6 & 0.05 \\
&                           &                           &                          & 2 & 80.8 & 0.07 \\
&                           &                           &                          & 5 & 88.2 & 0.10 \\
&                           &                           &                          & 10 & {92.2} & {0.13} \\
\cline{4-7}

&                           &                           & \multirow{4}{*}{Holo1-7B} & 1 & 55.4 & 0.02 \\
&                           &                           &                          & 2 & 66.3 & 0.03 \\
&                           &                           &                          & 5 & 75.7 & 0.05 \\
&                           &                           &                          & 10 & {80.4} & {0.06} \\

\hline

Operator & -- & -- & -- & -- & 87.0* & -- \\
Mariner  & -- & -- & -- & -- & 83.5* & -- \\
BrowserUse  & -- & -- & -- & -- & 89.1* & -- \\
\bottomrule
\end{tabular}

\end{table}

\subsection{WebVoyager Results}

\paragraph{Pareto-Optimality of \surferh+\hfamily}
Figure~\ref{fig:pareto-agents-retries} displays the performance of \surferh~for different policy modules (\hthreeb~or \hsevenb) and a set validator (GPT-4o). The Localizer selection is reported in Table~\ref{table:pareto-numbers}, along with exact benchmark numbers. Overall, \surferh+\hfamily~agents sit on the Pareto front, both at the 3B and 7B parameter models, and for all attempt values. This guarantees that using any of our models is always optimal. 

\surferh+\hsevenb~achieves a score of 92.2\% after 10 attempts, on par with \surferh+GPT-4.1 at 92.0\%, at a fraction of the cost (\$0.13/task \textit{vs} \$0.54/task). Both \hthreeb~and \hsevenb~outperform Operator and Mariner after just~5 attempts, and match Browser-Use's performance after 10 attempts. Powering \surferh~with either Gemini-2.0-Flash or Qwen2.5-VL-7B-Instruct leads to comparable performance (76.9\% and 78.2\%, respectively, after 10 attempts). Using Qwen2.5-VL-32B-Instruct boosts the score to 85.9\%, a few points below \hthreeb~(89.7\%).

\paragraph{\hfamily~as a Validator}
As our training mixture includes validation data, we also benchmark \surferh~using \hfamily~for the validator, focusing on Pareto-optimal agents, and compiling scores in \ref{table:pareto-numbers}.
This reduces the run cost, but the benchmark score drops by 16 percentage points and 12 percentage points at the 3B and 7B scales, respectively, after 10 attempts.
These fully \hfamily-powered agents remain close to the Pareto front, but the performance decline, in particular its reduction from 3B to 7B, is significant. This suggests that validation is harder than policy making or localization, or perhaps that it requires more cognitive power, and thus larger models. 

\paragraph{Performance on Unseen Tasks}
\label{sec:generalization}
We investigate how training on agent trajectories affects performance, depending on whether the evaluation tasks overlap with those seen during training. To this end, we create a version of our training data mixture that contains agent traces obtained from the WebVoyagerExtended task set only, and use it to train \hsevenb-WVE, an alternative to \hsevenb. For this experiment, \surferh~is allowed 10 attempts and uses GPT-4o as a Validator.

As shown in Table \ref{tab:gen}, \hsevenb-WVE yields a 9.5 percentage points boost over its foundation Qwen2.5-VL-7B-Instruct. The performance difference between \hsevenb-WVE and \hsevenb~(4.5 percentage points) illustrates the added benefit of in-domain experience. 
These results highlight the dual benefits of targeted (in-domain) fine-tuning and broad (cross-domain) exploration. When incorporated into the training of the policy model, these strategies substantially enhance agent performance, laying a foundation for reinforcement learning driven by large-scale agent executions.

\section{Conclusion} \label{sec:conclusion}

\surferh~and \hfamily~exemplify how powerful and cost-efficient web agents can be constructed on top of foundation models by tightly integrating modular architecture, targeted training strategies, and a rich and diverse data mixture.
By directly interacting with the web through the browser GUI, \surferh~offers a broad applicability across real-world tasks without the need for domain-specific integrations.

The introduction of the \hfamily~models, highly specialized vision models
trained to serve as \surferh~modules, enables both high performance and cost-effective deployment.
\hfamily~achieves state-of-the-art on both well-established and the newly introduced \webclick~localization benchmarks. 
Integrated into \surferh, these models lead to Pareto-optimal performance on the WebVoyager agentic benchmark. 
We hope that the release of our model weights and benchmark data will help catalyze future advances in agent research.

\bibliographystyle{abbrv}
\bibliography{refs}

\begin{thebibliography}{10}

\bibitem{anthropic2024claude3}
Anthropic.
\newblock Introducing computer use, a new claude 3.5 sonnet, and claude 3.5 haiku.
\newblock \url{https://www.anthropic.com/news/3-5-models-and-computer-use}, 2024.
\newblock Accessed: May 23, 2025.

\bibitem{bai2025qwen25vltechnicalreport}
S.~Bai, K.~Chen, X.~Liu, J.~Wang, W.~Ge, S.~Song, K.~Dang, P.~Wang, S.~Wang, J.~Tang, H.~Zhong, Y.~Zhu, M.~Yang, Z.~Li, J.~Wan, P.~Wang, W.~Ding, Z.~Fu, Y.~Xu, J.~Ye, X.~Zhang, T.~Xie, Z.~Cheng, H.~Zhang, Z.~Yang, H.~Xu, and J.~Lin.
\newblock {Qwen2.5-VL Technical Report}, 2025.

\bibitem{browseruse-tech-report-2024}
{Browser Use Team}.
\newblock Browser use: Sota technical report.
\newblock \url{https://browser-use.com/posts/sota-technical-report}, 2024.
\newblock Accessed May 23, 2025.

\bibitem{screenspot}
K.~Cheng, Q.~Sun, Y.~Chu, F.~Xu, L.~YanTao, J.~Zhang, and Z.~Wu.
\newblock {S}ee{C}lick: Harnessing {GUI} grounding for advanced visual {GUI} agents.
\newblock In {\em Proceedings of the 62nd Annual Meeting of the Association for Computational Linguistics (Volume 1: Long Papers)}, pages 9313--9332, Bangkok, Thailand, Aug. 2024. Association for Computational Linguistics.

\bibitem{dong2024xgrammar}
Y.~Dong, C.~F. Ruan, Y.~Cai, R.~Lai, Z.~Xu, Y.~Zhao, and T.~Chen.
\newblock Xgrammar: Flexible and efficient structured generation engine for large language models.
\newblock {\em arXiv preprint arXiv:2411.15100}, 2024.

\bibitem{gemini2025cost}
{Google Deepmind}.
\newblock {Gemini 2.0 Flash Docs}.
\newblock \url{https://ai.google.dev/gemini-api/docs/pricing#gemini-2.0-flash}.
\newblock Accessed: May 23, 2025.

\bibitem{google2024gemini}
{Google DeepMind}.
\newblock Introducing gemini 2.0: our new ai model for the agentic era.
\newblock \url{https://blog.google/technology/google-deepmind/google-gemini-ai-update-december-2024/}, December 2024.
\newblock Accessed: May 23, 2025.

\bibitem{google2024projectmariner}
{Google Deepmind}.
\newblock Project mariner: agents that can help you accomplish complex tasks.
\newblock \url{https://blog.google/technology/google-deepmind/google-gemini-ai-update-december-2024/#agents-for-developers}, 2024.
\newblock Accessed May 23, 2025.

\bibitem{gou2025uground}
B.~Gou, R.~Wang, B.~Zheng, Y.~Xie, C.~Chang, Y.~Shu, H.~Sun, and Y.~Su.
\newblock Navigating the digital world as humans do: Universal visual grounding for {GUI} agents.
\newblock In {\em The Thirteenth International Conference on Learning Representations}, 2025.

\bibitem{hartvigsen2022toxigenlargescalemachinegenerateddataset}
T.~Hartvigsen, S.~Gabriel, H.~Palangi, M.~Sap, D.~Ray, and E.~Kamar.
\newblock Toxigen: A large-scale machine-generated dataset for adversarial and implicit hate speech detection.
\newblock {\em arXiv preprint arXiv:2203.09509}, 2022.

\bibitem{webvoyager}
H.~He, W.~Yao, K.~Ma, W.~Yu, Y.~Dai, H.~Zhang, Z.~Lan, and D.~Yu.
\newblock {W}eb{V}oyager: Building an end-to-end web agent with large multimodal models.
\newblock In L.-W. Ku, A.~Martins, and V.~Srikumar, editors, {\em Proceedings of the 62nd Annual Meeting of the Association for Computational Linguistics (Volume 1: Long Papers)}, pages 6864--6890, Bangkok, Thailand, Aug. 2024. Association for Computational Linguistics.

\bibitem{GPT4-o-passes-bar-exam}
D.~M. Katz, M.~J. Bommarito, S.~Gao, and P.~Arredondo.
\newblock Gpt-4 passes the bar exam.
\newblock {\em Philosophical Transactions of the Royal Society A}, 2024.

\bibitem{koh2024visualwebarenaevaluatingmultimodalagents}
J.~Y. Koh, R.~Lo, L.~Jang, V.~Duvvur, M.~C. Lim, P.-Y. Huang, G.~Neubig, S.~Zhou, R.~Salakhutdinov, and D.~Fried.
\newblock Visualwebarena: Evaluating multimodal agents on realistic visual web tasks.
\newblock {\em arXiv preprint arXiv:2401.13649}, 2024.

\bibitem{laurençon2024matters}
H.~Laurençon, L.~Tronchon, M.~Cord, and V.~Sanh.
\newblock What matters when building vision-language models?
\newblock {\em arXiv preprint arXiv:2405.02246}, 2024.

\bibitem{lavague}
LaVagueAI.
\newblock Lavague: Web agent framework for builders.
\newblock \url{https://docs.lavague.ai/en/latest/}, 2025.
\newblock Accessed: May 23, 2025.

\bibitem{screenspot-pro}
K.~Li, Z.~Meng, H.~Lin, Z.~Luo, Y.~Tian, J.~Ma, Z.~Huang, and T.-S. Chua.
\newblock {ScreenSpot-Pro: GUI Grounding for Professional High-Resolution Computer Use}.
\newblock {\em arXiv preprint arXiv:2504.07981}, 2025.

\bibitem{browser_use2024}
M.~Müller and G.~Žunič.
\newblock Browser use: Enable ai to control your browser.
\newblock \url{https://github.com/browser-use/browser-use}, 2024.

\bibitem{nakano2022webgptbrowserassistedquestionansweringhuman}
R.~Nakano, J.~Hilton, S.~Balaji, J.~Wu, L.~Ouyang, C.~Kim, C.~Hesse, S.~Jain, V.~Kosaraju, W.~Saunders, X.~Jiang, K.~Cobbe, T.~Eloundou, G.~Krueger, K.~Button, M.~Knight, B.~Chess, and J.~Schulman.
\newblock {WebGPT: Browser-assisted question-answering with human feedback}.
\newblock {\em arXiv preprint arXiv:2112.09332}, 2022.

\bibitem{gpt41-2025cost}
{OpenAI}.
\newblock Gpt-4.1 documentation.
\newblock \url{https://platform.openai.com/docs/models/gpt-4.1}.
\newblock Accessed: May 23, 2025.

\bibitem{gpt41-mini-2025cost}
{OpenAI}.
\newblock Gpt-4.1-mini documentation.
\newblock \url{https://platform.openai.com/docs/models/gpt-4.1-mini}.
\newblock Accessed: May 23, 2025.

\bibitem{gpt4o2025cost}
{OpenAI}.
\newblock Gpt-4o documentation.
\newblock \url{https://platform.openai.com/docs/models/gpt-4o}.
\newblock Accessed: May 23, 2025.

\bibitem{gpt4o-mini2025cost}
{OpenAI}.
\newblock Gpt-4o-mini documentation.
\newblock \url{https://platform.openai.com/docs/models/gpt-4o-mini}.
\newblock Accessed: May 23, 2025.

\bibitem{openai2025operator}
{OpenAI}.
\newblock https://openai.com/index/introducing-operator/.
\newblock OpenAI Blog, Jan. 2025.

\bibitem{openai2025deepresearch}
{OpenAI}.
\newblock Introducing deep research.
\newblock \url{https://openai.com/index/introducing-deep-research/}, Feb. 2025.

\bibitem{gpt4}
{OpenAI \textit{et al.}}
\newblock {GPT-4 Technical Report}, 2024.

\bibitem{perplexity2025deepsearch}
{Perplexity Team}.
\newblock Introducing perplexity deep research.
\newblock \url{https://www.perplexity.ai/hub/blog/introducing-perplexity-deep-research}, 2025.
\newblock Accessed May 23, 2025.

\bibitem{qin2025ui}
Y.~Qin, Y.~Ye, J.~Fang, H.~Wang, S.~Liang, S.~Tian, J.~Zhang, J.~Li, Y.~Li, S.~Huang, et~al.
\newblock Ui-tars: Pioneering automated gui interaction with native agents.
\newblock {\em arXiv preprint arXiv:2501.12326}, 2025.

\bibitem{rawles2025androidworld}
C.~Rawles, S.~Clinckemaillie, Y.~Chang, J.~Waltz, G.~Lau, M.~Fair, A.~Li, W.~E. Bishop, W.~Li, F.~Campbell-Ajala, D.~K. Toyama, R.~J. Berry, D.~Tyamagundlu, T.~P. Lillicrap, and O.~Riva.
\newblock Androidworld: A dynamic benchmarking environment for autonomous agents.
\newblock In {\em The Thirteenth International Conference on Learning Representations}, 2025.

\bibitem{replit2022ghostwriter}
{Replit Team}.
\newblock Meet replit ghostwriter, your partner in code.
\newblock \url{https://blog.replit.com/ghostwriter}, October 2022.
\newblock Accessed: May 23, 2025.

\bibitem{rohrbach-etal-2018-object}
A.~Rohrbach, L.~A. Hendricks, K.~Burns, T.~Darrell, and K.~Saenko.
\newblock Object hallucination in image captioning.
\newblock In E.~Riloff, D.~Chiang, J.~Hockenmaier, and J.~Tsujii, editors, {\em Proceedings of the 2018 Conference on Empirical Methods in Natural Language Processing}, pages 4035--4045, Brussels, Belgium, Oct.-Nov. 2018. Association for Computational Linguistics.

\bibitem{schick2023toolformer}
T.~Schick, J.~Dwivedi-Yu, R.~Dess{\`\i}, R.~Raileanu, M.~Lomeli, L.~Zettlemoyer, N.~Cancedda, and T.~Scialom.
\newblock Toolformer: Language models can teach themselves to use tools.
\newblock {\em arXiv:2302.04761}, 2023.

\bibitem{shi2024languagemodelssolveolympiad}
Q.~Shi, M.~Tang, K.~Narasimhan, and S.~Yao.
\newblock Can language models solve olympiad programming?
\newblock {\em arXiv preprint arXiv:2404.10952}, 2024.

\bibitem{wei2022chain}
J.~Wei, X.~Wang, D.~Schuurmans, M.~Bosma, F.~Xia, E.~Chi, Q.~V. Le, D.~Zhou, et~al.
\newblock Chain-of-thought prompting elicits reasoning in large language models.
\newblock {\em Advances in neural information processing systems}, 35:24824--24837, 2022.

\bibitem{screenspot-v2}
Z.~Wu, Z.~Wu, F.~Xu, Y.~Wang, Q.~Sun, C.~Jia, K.~Cheng, Z.~Ding, L.~Chen, P.~P. Liang, and Y.~Qiao.
\newblock {OS}-{ATLAS}: Foundation action model for generalist {GUI} agents.
\newblock In {\em The Thirteenth International Conference on Learning Representations}, 2025.

\bibitem{xiao-wang-2021-hallucination}
Y.~Xiao and W.~Y. Wang.
\newblock On hallucination and predictive uncertainty in conditional language generation.
\newblock In P.~Merlo, J.~Tiedemann, and R.~Tsarfaty, editors, {\em Proceedings of the 16th Conference of the European Chapter of the Association for Computational Linguistics: Main Volume}, pages 2734--2744, Online, Apr. 2021. Association for Computational Linguistics.

\bibitem{xu2025theagentcompanybenchmarkingllmagents}
F.~F. Xu, Y.~Song, B.~Li, Y.~Tang, K.~Jain, M.~Bao, Z.~Z. Wang, X.~Zhou, Z.~Guo, M.~Cao, M.~Yang, H.~Y. Lu, A.~Martin, Z.~Su, L.~Maben, R.~Mehta, W.~Chi, L.~Jang, Y.~Xie, S.~Zhou, and G.~Neubig.
\newblock Theagentcompany: Benchmarking llm agents on consequential real world tasks, 2025.

\bibitem{yang2023setofmarkpromptingunleashesextraordinary}
J.~Yang, H.~Zhang, F.~Li, X.~Zou, C.~Li, and J.~Gao.
\newblock Set-of-mark prompting unleashes extraordinary visual grounding in gpt-4v.
\newblock {\em arXiv preprint arXiv:2310.11441}, 2023.

\bibitem{yao2023reactsynergizingreasoningacting}
S.~Yao, J.~Zhao, D.~Yu, N.~Du, I.~Shafran, K.~Narasimhan, and Y.~Cao.
\newblock {ReAct: Synergizing Reasoning and Acting in Language Models}.
\newblock {\em arXiv preprint arXiv:2210.03629}, 2023.

\bibitem{zhou2024webarena}
S.~Zhou, F.~F. Xu, H.~Zhu, X.~Zhou, R.~Lo, A.~Sridhar, X.~Cheng, T.~Ou, Y.~Bisk, D.~Fried, U.~Alon, and G.~Neubig.
\newblock {WebArena}: A realistic web environment for building autonomous agents.
\newblock {\em ICLR}, 2024.

\end{thebibliography}

\end{document}